# TRACE: Expert-Aligned ECG Representation Learning with Rigorous Benchmarking and Real-World Validation in Acute Cardiac Care

Lovely Yeswanth Panchumarthi [1], Andrew Lu [2], Saurabh Kataria [2], Delgersuren Bold [2], Minxiao Wang[2], Runze Yan[2], Patricia Dykes[2], Brian J. Gow [3], Tom J. Pollard[3], Jessica K. Zègre-Hemsey [4], Dillon J. Dzikowicz[5], Lekshmi Kumar [6], Xiao Hu[2], Ran Xiao [2]

[1] Department of Computer Science, Emory University, GA, USA

[2] School of Nursing, Emory University, GA, USA

[3] Institute for Medical Engineering and Science, Massachusetts Institute of Technology, MA, USA

[4] School of Nursing, University of North Carolina at Chapel Hill, NC, USA

[5] School of Nursing, University of Rochester, Rochester, NY, USA

[6] School of Medicine, Emory University, GA, USA

***Correspondence**

Ran Xiao, PhD, School of Nursing, Emory University, GA, USA. (ran.xiao@emory.edu)

## Abstract

TRACE (Text-Reinforced Analysis of Cardio ECGs) is a multimodal electrocardiogram (ECG) representation model that learns clinically grounded signal embeddings for downstream cardiac classification. It is designed to address the limitations of existing CLIP-style training, which often struggles with noisy clinical text and fails to leverage the complementary strengths of unimodal (from ECG) and cross-modal (between ECG and matched cardiologist reports) learning. To bridge this gap, we propose a hybrid architecture that jointly learns unimodal and cross-modal representations via uncertainty-weighted multi-task learning while utilizing an LLM-based pipeline to extract high-fidelity findings from cardiologist reports. We evaluate TRACE across a spectrum of clinical urgency, establishing robust performance on public benchmarks for arrhythmia classification and structural abnormalities relative to existing unimodal and multimodal ECG models. To demonstrate real-world utility, we further validate the model on acute coronary occlusion (ACO), where the prevailing ST-elevation criteria miss 25-34% of true occlusions. Utilizing a large private ACO dataset with expert-annotated ground truth, TRACE significantly outperforms real-world clinical practice, yielding a 19.0% increase in sensitivity or a 62.6% reduction in false positive rates at the clinical baseline. This extensive evaluation confirms that TRACE delivers both strong performance on benchmark tasks and tangible clinical impact in the most acute, high-risk cardiac scenarios.

## 1. Introduction

The 12-lead electrocardiogram (ECG) is among the most widely deployed diagnostic tools in clinical cardiology, essential for detecting arrhythmia, ischemic events, and structural abnormalities. According to the WHO, cardiovascular diseases accounting for approximately 32% of global mortality (World Health Organization, 2025), accurate and timely ECG interpretation directly impacts patient outcomes (Byrne et al., 2023; Rao et al., 2025). However, expert interpretation remains time-consuming, subject to inter-observer variability (Salerno et al., 2003), and constrained by a global shortage of trained providers (Maddox Thomas et al., 2024). These challenges motivate the development of ECG foundation models capable of transferring to diverse downstream tasks without task-specific annotations.

Recent efforts in ECG foundation models follow two directions: unimodal and multimodal approaches. Unimodal methods such as ST-MEM (Na et al., 2024), ECG-FM (McKeen et al., 2025), ECG-MAE (Hu et al., 2023) and CREMA (Song et al., 2024) learn representations through masked reconstruction or contrastive learning on signal augmentations. While effective at capturing morphological patterns, these approaches operate without access to clinical semantics, limiting their ability to encode the clinically meaningful distinctions that depend on expert interpretation of the signal. Multimodal methods such as MERL (Liu et al., 2024) address this gap by aligning ECG embeddings with paired clinical reports via contrastive learning, enabling zero-shot classification through text prompts.

Despite progress, two fundamental limitations persist: 1) Noisy text supervision in multimodal learning. Cardiologist reports in large-scale datasets such as MIMIC-IV-ECG contain heterogeneous content beyond ECG-specific observations: medication lists, prior imaging findings, patient history, and clinical context unrelated to the current tracing. When models align ECG signals with such noisy text, learned representations conflate signal-specific patterns with extraneous information, degrading downstream performance (Song et al., 2023). No prior work has systematically addressed the quality of text supervision itself for cardiologist reports, which encapsulate high-fidelity, domain-specific insights from expert clinicians. 2) Lack of complementary learning objectives. Existing multimodal methods rely solely on contrastive alignment, capturing global semantic relationships but missing fine-grained signal structure. Conversely, reconstruction-based methods learn local morphological features but lack semantic grounding. These objectives are complementary, yet combining them is non-trivial due to fundamental scale mismatch (uncertainty-weighted multi-task loss). Naive combination causes one objective to dominate optimization.

Beyond these two directions, a rapidly emerging line of work explores general-purpose multimodal time-series reasoning, in which large language models jointly interpret physiological signals and clinical text through natural-language prompting. Models such as QoQ-Med (Dai et al., 2026), OpenTSLM (Langer et al., 2025), and the approach of Parker et al. (Parker et al., 2025a) operate over diverse multivariate time-series and generate open-ended, free-text reasoning via generative decoders, and are typically evaluated on question-answering benchmarks requiring natural-language justification. TRACE is deliberately scoped to a different and complementary objective: it is a domain-specialized representation model for the 12-lead, 10-second clinical ECG, the most widely deployed cardiac diagnostic signal that produces structured predictions for high-stakes, time-sensitive triage (e.g., acute coronary occlusion (ACO) detection) rather than open-ended narrative reasoning. This specialization is precisely what enables the ECG-specific report curation and clinically grounded validation that underpin our contributions; we therefore regard general-purpose reasoning systems as a complementary rather than competing paradigm.

In this work, we present TRACE, a multimodal ECG representation model that addresses both limitations. First, we develop an LLM-based curation pipeline to extract ECG-specific findings from MIMIC-IV cardiologist reports, filtering lacking interpretable cardiac content. Second, we propose a hybrid architecture that jointly optimizes masked autoencoder reconstruction (MAE) and contrastive multimodal alignment (CMA) within a shared encoder. To handle the scale mismatch between objectives, we employ uncertainty-weighted multi-task learning that automatically balances loss contributions based on learned task uncertainties. We rigorously evaluate TRACE across a wide spectrum of cardiac tasks using common public benchmarks, demonstrating strong performance in both structural abnormality detection and arrhythmia classification against existing literature. Most significantly, we validate the model in a clinically tangible setting using a large-scale private dataset for ACO, a condition severely underserved by current AI-based clinical decision-making. This comprehensive evaluation addresses the critical gap between algorithmic development and medical implementation, proving that TRACE provides a viable, high-impact solution for high-stakes cardiac triage. While the individual components we build on masked reconstruction, contrastive alignment, and uncertainty weighting are established techniques, our contribution is domain-strategic: it lies in how they are curated, integrated, and clinically validated for ECG interpretation rather than in isolated algorithmic invention.

## 2. Methods

### 2.1 Source Dataset and LLM-Based Curation

#### *2.1.1 MIMIC-IV ECG Dataset*

We utilize MIMIC-IV-ECG (Gow et al., 2023), a large-scale collection of clinical 12-lead electrocardiograms recorded at Beth Israel Deaconess Medical Center from 2008 to 2019. Each recording consists of 12 simultaneous leads sampled at 500 Hz for 10 seconds, yielding 5,000 time points per lead. The dataset contains 800,035 ECG waveforms and 623,566 cardiologist reports. After linking waveforms with their corresponding reports and excluding records with missing data, we obtain 609,272 ECG-report pairs covering 105,293 unique patients.

#### *2.1.2 Report Curation Pipeline*

Cardiologist reports in MIMIC-IV-ECG contain heterogeneous content extending beyond ECG-specific observations: medication changes, prior imaging findings, patient history, and clinical context not inferable from the current ECG tracing. Aligning ECG embeddings with such noisy text causes learned representations to conflate signal-specific patterns with extraneous information. To address this, we develop a three-stage curation pipeline using Qwen3-32B, as illustrated in Figure 1(a).

Stage (i): **Input Filtering.** Raw reports first pass through a keyword-based filter that removes records containing only non-informative references (e.g., "Refer to previous report"). These records lack interpretable ECG content and provide no meaningful supervision signal. Stage (ii): **LLM-Based Extraction.** Remaining reports are processed by Qwen3-32B using a structured prompt with two components: a cardiologist persona prompt that instructs the model to decompose findings into base diagnostic terms, and decomposition rules that constrain extraction to retain only ECG-attributable observations (rhythm descriptions, axis deviations, ST-T changes) while removing medication lists, imaging references, and historical context. The complete prompt is provided in Appendix B. Stage (iii): **Terminology Standardization.** Extracted findings undergo terminology expansion using a clinical normalization dictionary that maps abbreviations to standardized terms (e.g., "RVH" → "right ventricular hypertrophy"). This ensures consistent vocabulary while preserving clinical semantics. The dictionary, following SCP coding guidelines consistent with prior work (Liu et al., 2024), is provided in Appendix C. The pipeline removes 115,020 samples (19%), yielding 494,252 high-quality ECG-report pairs where text supervision provides meaningful signal for the ECG encoder. The removed reports fall into two groups: those flagged by the Stage (i) keyword filter as containing only non-informative references (e.g., “Refer to previous report”), and those from which Qwen3-32B extracted no ECG-attributable finding typically reports dominated by medications, prior imaging, or patient history rather than the current tracing. The retained pairs instead carry the observations a cardiologist reads directly from the signal rhythm, axis, conduction abnormalities, and ST–T morphology making them high-fidelity supervision targets that keep the contrastive objective from binding ECG embeddings to information not inferable from the tracing. The pipeline was run entirely on-premise from open-source model weights with no external API; full deployment and compute details are provided in Appendix B.

**2.2 Model Architecture**

*2.2.1 Overview*

TRACE employs a hybrid architecture that jointly learns unimodal signal representations through masked autoencoder reconstruction and cross-modal representations through contrastive alignment with clinical text, as detailed in Figure 1 (b). Given an ECG-report pair $(\boldsymbol{X}, \boldsymbol{T})$ where $\boldsymbol{X} \in \mathbb{R}^{12\times5000}$ denotes the 12-lead signal and $\boldsymbol{T}$ denotes the clinical report, TRACE learns representations through two complementary objectives. The masked autoencoder path randomly occludes a subset of signal patches and trains the encoder to produce representations sufficient for reconstructing the masked content, encouraging fine-grained morphological understanding. The contrastive path aligns the global ECG representation with the corresponding text embedding, grounding signal features in clinical semantics. Both paths share the ECG encoder, enabling the model to learn representations that capture both local signal structure and global clinical meaning.

*2.2.2 ECG Encoder*

The ECG encoder $f_\theta$ maps the input signal to a sequence of contextualized representations. We first partition the 12-lead ECG into $N$ non-overlapping temporal patches, where each patch spans $P$ time points. For a 10-second recording at 500 Hz with patch size $P = 50$, this yields $N = 12 \times 100 = 1200$ patches. A patch embedding function $\phi: \mathbb{R}^P \rightarrow \mathbb{R}^d$ projects each patch to the latent dimension $d$. Following standard practice in vision transformers [dosovitskiy2020image], we prepend a learnable classification token $\boldsymbol{z}_{\text{cls}}$ and add positional embeddings $\boldsymbol{E}_{\text{pos}}$ encoding both lead identity and temporal position:

$$\boldsymbol{Z}^{(0)} = [\boldsymbol{z}_{\mathrm{cls}}; \phi(\boldsymbol{x}_1); \dots; \phi(\boldsymbol{x}_N)] + \boldsymbol{E}_{\mathrm{pos}}$$

The sequence passes through $L$ transformer blocks, each applying multi-head self-attention followed by a feed-forward network with pre-normalization:

$$\boldsymbol{Z}^{(l)} = \boldsymbol{Z}^{(l-1)} + \mathrm{FFN}\big(\mathrm{LN}(\boldsymbol{Z}^{(l-1)} + \mathrm{MHSA}(\mathrm{LN}(\boldsymbol{Z}^{(l-1)})))\big)$$

The final classification token $\boldsymbol{z}_{\mathrm{cls}}^{(L)}$ serves as the global ECG representation for contrastive alignment, while patch representations are used for reconstruction.

*2.2.3 Masked Autoencoder Reconstruction*

For the reconstruction path, we follow the asymmetric encoder-decoder design. During training, we sample a binary mask $m \in \{0,1\}^N$ where each entry is independently drawn with masking probability $\rho$. The encoder operates only on visible patches $\mathcal{V} = \{i : m_i = 0\}$, reducing computational cost proportional to $(1-\rho)^2$.

The decoder $g_\psi$ reconstructs masked patches from encoded visible representations. Let $\boldsymbol{Z}_\mathcal{V}$ denote encoder outputs for visible patches. The decoder input is formed by inserting learnable mask tokens $\boldsymbol{e}_{\mathrm{mask}}$ at masked positions and adding decoder-specific positional embeddings:

$$\widehat{\boldsymbol{X}} = g_\psi\big(\mathrm{Concat}(\boldsymbol{Z}_\mathcal{V}, \boldsymbol{e}_{\mathrm{mask}}) + \boldsymbol{E}'_{\mathrm{pos}}\big)$$

The decoder is intentionally lightweight—shallower and narrower than the encoder—forcing the encoder to learn rich representations rather than deferring reconstruction to decoder capacity.

*2.2.4 Text Encoder and Projection*

For text encoding, we employ a pretrained biomedical language model $h_\phi$ that maps tokenized reports to dense representations. Given report $\boldsymbol{T}$, we obtain the text embedding as the pooled output: $\boldsymbol{h}_{\text{text}} = h_\phi(\boldsymbol{T})$. The text encoder remains frozen during training to preserve pretrained clinical knowledge.

To align ECG and text representations in a shared semantic space, we apply modality-specific projectors $\pi_e$ and $\pi_t$ for ECG and text respectively. The aligned representations are $\ell_2$-normalized to the unit hypersphere:

$$\boldsymbol{z}^e = \frac{\pi_e(\boldsymbol{z}_{\text{cls}}^{(L)})}{\| \pi_e(\boldsymbol{z}_{\text{cls}}^{(L)}) \|_2}, \quad \boldsymbol{z}^t = \frac{\pi_t(\boldsymbol{h}_{\text{text}})}{\| \pi_t(\boldsymbol{h}_{\text{text}}) \|_2}$$

Both projectors are two-layer MLPs mapping to a shared $d'$-dimensional space, ensuring that similarity is measured by cosine distance.

**2.3 Training Objectives**

TRACE optimizes two complementary objectives: masked autoencoder reconstruction for learning fine-grained signal features, and contrastive multimodal alignment for grounding representations in clinical semantics.

*2.3.1 Masked Autoencoder Reconstruction Loss*

The reconstruction objective encourages the model to learn meaningful signal representations by predicting masked patch values from visible context. We compute the loss only over masked positions using the L1 norm:

$$\mathcal{L}_{\text{MAE}} = \frac{1}{|\mathcal{M}|} \sum_{i \in \mathcal{M}} | \boldsymbol{x}_i - \hat{\boldsymbol{x}}_i |$$

where $\boldsymbol{x}_i$ and $\hat{\boldsymbol{x}}_i$ denote the original and reconstructed patch values respectively, and $\mathcal{M}$ is the set of masked patch indices.

*2.3.2 Contrastive Multimodal Alignment Loss*

The alignment objective learns a shared embedding space where matched ECG-report pairs are close while unmatched pairs are distant. We employ symmetric InfoNCE loss (Oord et al., 2018) over the batch:

$$\mathcal{L}_{\text{CMA}} = -\frac{1}{2N}\sum_{i=1}^{N}[\log\frac{\exp(\boldsymbol{z}_i^e \cdot \boldsymbol{z}_i^t/\tau)}{\sum_{j=1}^{N}\exp(\boldsymbol{z}_i^e \cdot \boldsymbol{z}_j^t/\tau)} + \log\frac{\exp(\boldsymbol{z}_i^t \cdot \boldsymbol{z}_i^e/\tau)}{\sum_{j=1}^{N}\exp(\boldsymbol{z}_i^t \cdot \boldsymbol{z}_j^e/\tau)}]$$

where $N$ is the batch size and $\tau = 0.07$ is a temperature parameter controlling the sharpness of the similarity distribution.

*2.3.3 Uncertainty-Weighted Multi-Task Loss*

The reconstruction and alignment objectives operate at fundamentally different scales: MAE losses range from 0.02 to 0.05, while contrastive losses range from 4 to 8. Naive summation or fixed weighting schemes fail to balance these objectives effectively, often causing one task to dominate optimization.

We address this scale mismatch through uncertainty-weighted multi-task learning, which treats task-specific noise as a learnable quantity. Following [cipolla2018a], we parameterize the combined loss as:

$$\mathcal{L}_{\text{total}} = \frac{1}{2\sigma_1^2}\mathcal{L}_{\text{MAE}} + \frac{1}{2\sigma_2^2}\mathcal{L}_{\text{CMA}} + \log\sigma_1 + \log\sigma_2$$

where $\sigma_1^2$ and $\sigma_2^2$ are learned observation noise parameters for each task. The precision terms $1/\sigma^2$ downweight tasks with higher uncertainty, while the logarithmic regularizers prevent uncertainties from growing unboundedly. In practice, we learn the log-variances $s_i = \log\sigma_i^2$ directly for numerical stability.

### 2.4 Training Parameters

We pretrain TRACE on the curated MIMIC-IV dataset for 100 epochs using AdamW optimizer (Loshchilov & Hutter, 2017) with base learning rate $10^{-3}$, weight decay 0.05, and $(\beta_1, \beta_2) = (0.9, 0.95)$. The learning rate follows a cosine decay schedule with 40 warmup epochs. We use a batch size of 128 per GPU with mixed-precision training. The mask ratio is set to 0.75 for MAE pretraining. Training is conducted on NVIDIA L40S GPUs with approximately 48 hours for full convergence. We select the best checkpoint based on a weighted combination of validation reconstruction loss and contrastive alignment accuracy.

Our study builds on MERL (Liu et al., 2024), the multimodal ECG-text contrastive framework that first demonstrated zero-shot ECG classification from paired clinical reports. We adopted MERL's publicly released codebase as the starting point for our contrastive alignment branch and drew inspiration from its formulation of ECG-report alignment. TRACE nonetheless departs substantially from MERL in both objective and implementation: we introduce (i) an LLM-based report-curation pipeline that filters noisy text supervision prior to alignment, (ii) a hybrid objective coupling masked autoencoder reconstruction with contrastive alignment inside a shared encoder rather than contrastive alignment alone and (iii) uncertainty-weighted multi-task balancing to reconcile the two objectives' disparate scales. Consequently, while this work originated from MERL, the final architecture, training procedure, and codebase differ significantly. We thank the MERL authors for releasing their code, which facilitated this research.

## 2.5 Experiments

### *2.5.1 Benchmarking TRACE on Public Datasets*

We evaluate TRACE on three widely used public cardiac benchmark datasets, including PTB-XL (Wagner et al., 2020) (21,837 ECGs, three multi-label classification tasks), CPSC2018 (Liu et al., 2018) (6,877 ECGs, 9 rhythm classes), Chapman-Shaoxing-Ningbo (CSN) (Zheng et al., 2020) (45,152 ECGs, 11 rhythm annotations). Together, these datasets encompass a diverse range of cardiac etiologies, including structural abnormalities, acute myocardial ischemia, and cardiac arrhythmia. This makes them robust platform to assess the capability of TRACE to extract physiologically relevant features from raw ECG waveforms. Detailed dataset characteristics and task definitions are provided in Appendix A. By following the official splits in these datasets, we benchmark TRACE against prominent ECG foundation models in literature trained with unimodal self-supervised methods and multimodal ECG-text approaches, ensuring a transparent and reproducible performance comparison. The full list of ECG foundation models is provided in Table 1. In total, Table 1 compares TRACE against 15 published baselines spanning both paradigms: 11 unimodal models covering contrastive (SimCLR, MoCo-v3, TS-TCC, CLOCS, ASTCL), non-contrastive (BYOL, BarlowTwins, SimSiam), masked/generative (CRT, ST-MEM), and large-scale supervised (ECGFounder) pretraining and four multimodal ECG–text methods (MERL with ResNet and ViT backbones, K-MERL, and SuPreMe). To our knowledge this is among the most comprehensive baseline comparisons in the ECG foundation-model literature, all reported under identical linear-probing evaluation with AUROC on official splits.

We adopt two standard evaluation protocols. Linear probing freezes the pretrained encoder and trains only a linear classification head, directly measuring representation quality of TRACE without confounding from downstream optimization. Fine-tuning updates all parameters end-to-end with a reduced learning rate on the encoder, assessing the model's capacity to adapt learned features to specific tasks. For both protocols, we train downstream classifiers using AdamW with cosine learning rate decay. We report the area under the receiver operating characteristic curve (AUROC) as the primary metric, averaged over five random seeds with standard deviations, shown as error bars in Figure 2. All public downstream experiments use the official data splits where available.

*2.5.2 Clinical Evaluation on ACO Detection*

Accurate identification of ACO remains clinically challenging because roughly one in three patients with angiographically confirmed acute coronary occlusion do not meet traditional ST-elevation criteria at initial presentation and are instead classified as NSTEMI (Meyers et al., 2021; Ricci et al., 2025). As a result, reliance on the STEMI framework alone can delay reperfusion therapy, leading to prolonged ischemic time, increased infarct size, and worse clinical outcomes.

To address this diagnostic gap, we designed a clinically grounded evaluation framework to assess whether TRACE improves ACO detection beyond routine clinical practice. Specifically, we quantitatively compared TRACE against a real-world clinical practice performance baseline using two complementary operating-point analyses: (1) matching the false positive rate (FPR) of clinical practice and measuring the corresponding sensitivity gain achieved by TRACE, and (2) matching sensitivity (true positive rate, TPR) and evaluating the reduction in FPR. This bidirectional matching strategy provides a concrete and interpretable assessment of clinical utility, directly reflecting trade-offs between patient rule-in and rule-out encountered in real-world decision-making.

We also qualitatively evaluated TRACE to assess whether its predictions align with clinically meaningful patterns. We first explored low-dimensional visualization of learned representations using t-SNE to examine class separability. In addition, we computed saliency mapping (Simonyan et al., 2013) on randomly selected true-positive ACO cases to illustrate features driving correct model predictions. This approach computes the gradient of the model's loss with respect to the input ECG signal, providing a direct measure of how sensitive the model's prediction is to different section of input signal. To enhance interpretability, we applied min-max normalization to the saliency values across all leads and smoothed the data using a rolling average with a window of 10 time steps. These analyses aim to contextualize model behavior and enhance clinical interpretability, supporting trust and adoption in practice.

All clinical evaluations were performed using a large private ACO dataset derived from county-level emergency medical services (EMS) records linked with in-hospital electronic health record (EHR) data. Detailed dataset description is provided in Appendix A. Ground-truth ACO labels were expert-adjudicated through comprehensive chart reviews to mirror the diagnostic complexity of real-world practice. We benchmarked TRACE against a clinical baseline following standard AHA/ACC STEMI guidelines (Rao et al., 2025). Importantly, this clinical practice baseline is not a simulated decision rule but the triage decision the treating team actually made at presentation: whether the patient received a catheterization-lab STEMI activation based on the ECG read under conventional ST-elevation criteria, extracted from linked activation records. Against the expert-adjudicated ACO ground truth, an activation with an angiographically confirmed occlusion counts as a true positive, an activation without confirmed occlusion as a false positive, and a confirmed occlusion that received no activation as a false negative. These definitions yield the clinical operating point (sensitivity 0.781, false positive rate 0.503) against which TRACE is compared in Figure 4. TRACE's ACO performance is itself reported with 10-fold cross-validation, each fold using a 70/10/20 train/validation/test split. By integrating operating-point quantitative metrics with clinically grounded qualitative analysis, we provide a robust assessment of TRACE's efficacy as a decision-support tool for enhancing ACO detection during the high-stakes initial phases of patient care.

## 3. Results

### 3.1 TRACE Benchmark Performance

Figure 2 presents TRACE's performance across all downstream benchmarks using our best-performing configuration: TRACE (CMA+MAE) pretraining with 75% mask ratio and uncertainty-weighted loss balancing. A systematic ablation analysis, provided in Appendix D, validates these design choices by isolating the contributions of individual components, including the relative impact of MAE versus CMA, optimal masking ratios, loss balancing strategies, and the selection of the text encoder. Notably, our domain-specific MedCPT encoder outperforms a nearly 300 × larger general-purpose LLM (Qwen3-32B) as the text backbone (Table A.4), indicating that alignment-compatible pretraining matters more than raw model scale for ECG–text supervision. TRACE achieves competitive performance under linear probing, with only marginal improvements observed following full fine-tuning across most downstream benchmarks. This narrow performance gap suggests that the pretrained representations effectively encode high-fidelity, task-relevant features, reducing the need for extensive task-specific adaptation. Performance peaks on the PTB-XL rhythm classification subtask (96.0% AUROC under finetuning), followed by CSN (95.5% under finetuning), CPSC2018 (95.0% under finetuning), and the PTB-XL Superclass (93.3% under finetuning) and Subclass (92.6% under linear probing) tasks. Collectively, these results indicate that TRACE's hybrid pretraining objective effectively captures complementary signal-level morphology and high-level semantic information, yielding robust and transferable ECG representations.

**3.2 Comparison with Unimodal Self-Supervised Methods**

Table 1 compares TRACE against unimodal self-supervised learning (SSL) baselines. TRACE substantially outperforms the best unimodal method across all benchmarks. This performance gap is consistent with our hypothesis that naive signal-level augmentations commonly used to construct positive pairs in contrastive learning may distort clinically meaningful ECG morphology. In particular, methods such as cutout or temporal cropping may inadvertently mask diagnostic features, including ST-segment elevation or QRS complex width, causing models to become insensitive to precisely the patterns clinicians rely on for interpretation.

Generative objectives successfully avoid semantic distortion by learning low-level signal patterns (e.g., local waveform intensities), but they lack access to high-level diagnostic semantics. TRACE addresses both limitations: the MAE branch learns fine-grained morphology without corrupting input signals, while the CMA branch grounds these features in clinical knowledge from curated reports. This complementarity is quantified in our component ablation (Table A.1): MAE-only reconstruction on the same encoder reaches 0.8623 AUROC on PTB-XL Super and 0.7965 on ACO under linear probing; adding contrastive alignment raises these to 0.9264 and 0.8635 a +6.4-point gain on PTB-XL Super confirming that multimodal supervision, not architecture alone, drives the improvement over unimodal pretraining.

Among the unimodal baselines, the supervised ECGFounder is by far the strongest, which we attribute to scale: it is trained on over 10 million labeled ECGs spanning 150 conditions roughly 20× the ~ 494K curated pairs used by TRACE. Two observations follow. First, despite this data disadvantage, TRACE still surpasses ECGFounder on all five public benchmarks, indicating that high-quality, curated multimodal supervision can be more sample-efficient than large-scale supervised pretraining. Second, ECGFounder trails the multimodal methods MERL and K-MERL on CPSC2018 and CSN, suggesting that supervision tied to a fixed label taxonomy transfers less readily to datasets with different annotation schemas, whereas contrastive alignment with free-text reports provides more flexible semantic grounding.

### 3.3 Comparison with Multimodal ECG-Text Methods

Existing multimodal methods rely primarily on contrastive alignment between ECG signals and clinical reports. Table 1 shows that TRACE outperforming K-MERL by 2.81 points and MERL with ResNet encoder by 3.97 points on PTBXL-Super benchmark. These improvements over MERL variants highlight the effectiveness of our hybrid architecture. While contrastive learning aligns global ECG representations with clinical semantics, it does not explicitly encourage the encoder to capture local morphological details essential for distinguishing subtle pathological patterns. TRACE's MAE objective addresses this gap by training the encoder to reconstruct masked signal segments, forcing it to learn fine-grained features that complement the high-level semantics learned through text alignment.

Additionally, our LLM-based report curation ensures that the contrastive objective operates on clean supervision signals, reducing noise introduced when models attempt to align ECG patterns with irrelevant clinical information, such as medication lists or historical imaging findings. The combination of architectural innovations and improved data quality accounts for TRACE's consistent performance gains across all evaluated benchmarks.

### 3.4 Clinical Evaluation

#### *3.4.1 TRACE Improves ACO Rule-in and Rule-Out*

TRACE demonstrated a strong overall performance with AUROC of 0.86 ± 0.01 under linear probing evaluation and 0.87 ± 0.02 under fine-tuning evaluation. We further evaluated the diagnostic performance of our TRACE model for ACO detection using ROC curve analysis, comparing it directly against an expert-chart-review-based clinical practice baseline (Figure 4). The clinical baseline (red dot), derived from the private ACO dataset with linked catheterization lab reports through rigorous expert chart review, operated at a sensitivity of 0.781 and a FPR of 0.503 (specificity = 0.497), reflecting the inherent challenges of routine ACO diagnosis. TRACE consistently demonstrates superior discriminative power over the established clinical baseline across the clinically relevant range of the ROC space. Specifically, at a matched sensitivity of 0.781, TRACE substantially reduced the FPR to 0.188 (purple dot), marking a 62.6% reduction in false positives compared to standard clinical practice. Conversely, when constrained to the clinical baseline's FPR of 0.503, the model achieved a sensitivity (TPR) of 0.929 (yellow dot), representing a 19.0% relative gain in detecting previously under-recognized occlusions.

In our ACO dataset, TRACE's performance gain translates to the identification of 90 additional ACO patients over six years. Retrospective chart review revealed that these patients were initially misclassified as less-urgent non-STE myocardial infarction (NSTEMI), resulting in a mean time-to-catheterization of 20.8 hours that is nearly 25 times longer than the 0.81 hours observed for those correctly identified as ACO. This diagnostic delay was associated with a 26.0% higher incidence of major adverse cardiac events (MACE). This gap mirrors national trends in the U.S. where 25–34% of NSTEMI cases involve undiagnosed coronary occlusions, while 14–36% of STEMI activations remain false positives (Larson et al., 2007; Ricci et al., 2025; Shamaki et al., 2024). Conversely, TRACE identified up to 1,639 patients over six years in the ACO dataset, who were previously referred for catheterization despite the absence of true occlusion. Such false activations impose a significant economic burden, costing upwards of $10,000 per case, while exposing patients to avoidable procedural risks (Degheim et al., 2019). Together, these diagnostic gains align with contemporary quality metrics that prioritize both rapid revascularization and high-value, safe clinical triage.

*3.4.2 Physiologic Grounding and Interpretability*

The t-SNE visualization of the learned ECG embeddings reveals a moderate, intrinsic separation between ACO and non-ACO samples (Figure 3a). Notably, this clustering emerged without task-specific supervised training, suggesting that the model's self-supervised objective captures fundamental ischemic features. This high-dimensional separation likely underpins the model's robust diagnostic performance and its potential generalizability across diverse cardiac pathologies, as the representations are not over-fitted to one specific task.

By utilizing lead-wise saliency maps (Figure 3b), where darker intensity indicates higher model importance, we observed that the model's attention spans multiple components of the cardiac cycle beyond just ST-segment used in the clinical guidelines. While ST-segment deviation and morphology contribute to model importance, substantial saliency is also observed over the QRS complex and T-wave, reflecting ischemia-related alterations in ventricular depolarization and repolarization. In particular, emphasis on terminal QRS morphology, R-wave progression, and repolarization abnormalities suggests sensitivity to conduction slowing, altered activation vectors, and early ischemic changes that often precede or occur without diagnostic ST elevation. The presence of salient regions in reciprocal and anatomically opposing leads further indicates that the model integrates global vectorial and reciprocal electrical patterns characteristic of acute coronary occlusion. Collectively, these findings reinforce the emerging paradigm that ACO is a complex pathophysiologic entity characterized by a constellation of subtle, multi-segment ECG abnormalities rather than a discrete ST-elevation phenotype (Karim et al., 2025; McLaren et al., 2024). This also explains the superior performance of TRACE, as its architecture is specifically designed to capture these nuanced features that are often overlooked by the conventional STE paradigm.

# 4. Discussion

## 4.1 Two paradigms for combining ECG and clinical text

Work combining ECG signals with clinical text falls into two paradigms that differ in when text is used and what the model outputs at inference. In the *text-as-supervision* paradigm, reports are consumed only during pretraining as alignment targets; at inference the model takes the ECG alone and emits structured outputs (class probabilities, calibrated risk scores) through a linear head or fine-tuning the dominant setting for clinical ECG classification, including MERL [liu2024zero], K-MERL (Liu et al., 2024), SuPreMe (Cai et al., 2025), C-MELT (Pham et al., 2024), ECG-FM (McKeen et al., 2025), and ECGFounder (Li et al., 2025). In the *text-as-input* paradigm, the ECG conditions a generative language model that produces free-text answers for question answering, report generation, and open-ended reasoning, as in QoQ-Med (Dai et al., 2026), OpenTSLM (Langer et al., 2025), TsLLM (Parker et al., 2025b), and ECG-specific systems such as ECG-Chat (Zhao et al., 2025) and MEIT (Wan et al., 2025). TRACE belongs squarely to the first paradigm: its LLM curation parallels the report-entity extraction of SuPreMe and K-MERL, and its hybrid masked-reconstruction-plus-contrastive objective parallels C-MELT.

## 4.2 Relation to general-purpose reasoning models

Because TRACE contains no generative decoder, it cannot emit free-text reasoning, and open-ended benchmarks such as ECG-QA (Oh et al., 2023) are not applicable without a language-generation head. Crucially, this is not a disadvantage for our target task: on closed-set classification the representation-learning paradigm remains stronger, and the reasoning-model papers report this themselves OpenTSLM (Langer et al., 2025) reports F1 $\approx 40$ on ECG-QA versus $\approx 76$ for a fixed-head classifier on the same labels, and TsLLM (Parker et al., 2025b) states its model is "not designed to surpass specialized models on traditional benchmarks." These systems position their contribution as additive open-ended capability, not a replacement for specialized classifiers. We therefore regard the two paradigms as complementary: reasoning models suit explanatory and exploratory use, whereas TRACE targets decision-ready predictions for high-stakes 12-lead ECG triage.

### 4.3 Relation to imaging-paired cross-modal ECG models

A related family of multimodal ECG models pairs the ECG with another *imaging* modality rather than text: MEDBind (Gao et al., 2024) unifies ECG with chest X-ray, the approach of Turgut et al. (Turgut et al., 2025) transfers information from cardiac magnetic resonance, and EchoingECG (Gao et al., 2025) aligns ECG with echocardiography. These are conceptually related but methodologically distinct from TRACE, which uses cardiologist report text as the pairing signal. The distinction is practically consequential: imaging modalities such as CMR and echocardiography are seldom available at the moment an ECG is acquired least of all in the prehospital and emergency settings where ACO triage occurs whereas paired report text is abundant and requires no co-registered imaging. A direct comparison against these imaging-paired models on shared PTB-XL tasks is a valuable direction for future work.

### 4.4 Concurrent self-supervised ECG models

Several concurrent efforts also release pretrained ECG encoders. HuBERT-ECG (Coppola et al., 2024) adapts masked-unit prediction to the ECG and targets a broad set of 164 diagnostic categories, while the DeepECG.ai platform of Nolin-Lapalme et al. (Nolin-Lapalme et al., 2026) scales self-supervised ECG analysis across large clinical cohorts. Because these models are developed and evaluated on cohorts, label taxonomies, and data splits that differ from the PTB-XL (Super/Sub/Rhythm), CPSC2018, and CSN configurations used in Table 1, a faithful numerical comparison would require reimplementing and re-evaluating each under identical protocols. We therefore note them as concurrent, complementary work rather than as directly comparable baselines.

### 4.5 Limitations

Several limitations temper our findings. First, the clinical validation relies on a single county-level EMS cohort with a closed, private ACO dataset; despite expert adjudication, external multi-site and prospective validation is needed before deployment, and the clinical practice baseline is retrospective rather than from a controlled trial. Second, TRACE is a representation model with structured outputs by design: this avoids the hallucination, calibration, and grounding concerns that lead generative ECG–language systems such as QoQ-Med (Dai et al., 2026) to carry explicit not-for-deployment statements, but it also means TRACE cannot produce the open-ended explanations or interactive reasoning those models target. Third, our multimodal supervision is text-only; TRACE does not integrate complementary imaging modalities (e.g., CMR, echocardiography) that could further enrich representations where such data are available. Finally, the model is specialized to the standard 12-lead, 10-second ECG and does not address other cardiac data types or acquisition settings. We view these as scope boundaries that motivate future work prospective evaluation, controlled integration of additional modalities, and potentially coupling the encoder to a reasoning module rather than as fundamental barriers.

## 5. Conclusion

We presented TRACE, a multimodal foundation model bridging ECG representation learning and clinical practice. By employing an LLM-based curation pipeline, we distilled noisy narratives into high-fidelity supervision signals grounded in expert expertise. Our hybrid architecture further optimizes for both fine-grained morphology and global semantics using an uncertainty-weighted loss to balance multi-scale objectives. Rigorous evaluation demonstrates that while TRACE maintains state-of-the-art performance on public benchmarks, it achieves a transformative impact in ACO detection, significantly outperforming real-world clinical baselines. These results establish a scalable pathway for expert-aligned, trustworthy AI to improve high-stakes triage and patient outcomes in acute cardiac care.

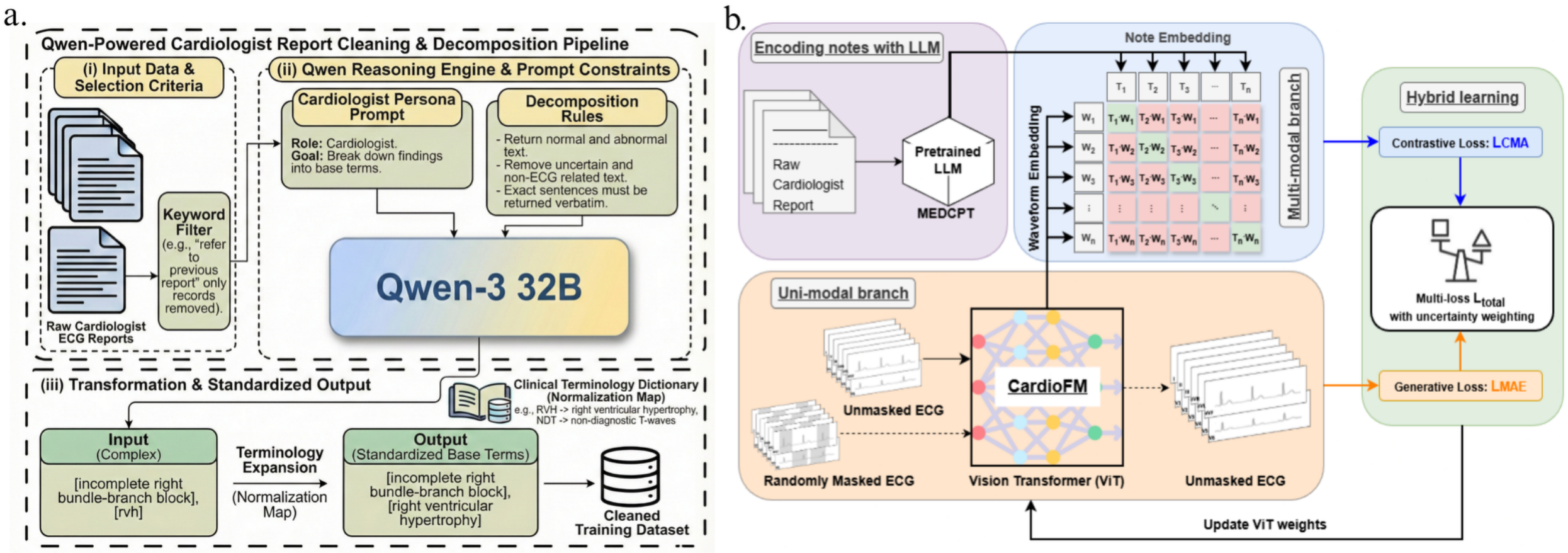

Figure 1. Overview of TRACE. (a) Qwen-3 32B-based pipeline for extracting ECG-specific findings from cardiologist reports, and (b) hybrid architecture combining masked reconstruction with contrastive ECG-text alignment. TRACE follows a two-phase workflow: (i) hybrid self-supervised pretraining that jointly optimizes masked autoencoder reconstruction (MAE) and contrastive multimodal alignment (CMA) under uncertainty-weighted loss balancing (panels a–b); and (ii) downstream evaluation, in which the shared ECG encoder's [CLS] token is passed to a linear classification head (linear probing) or fine-tuned end-to-end to produce task-specific predictions (e.g., multi-label AUROC on PTB-XL Super/Sub/Rhythm).

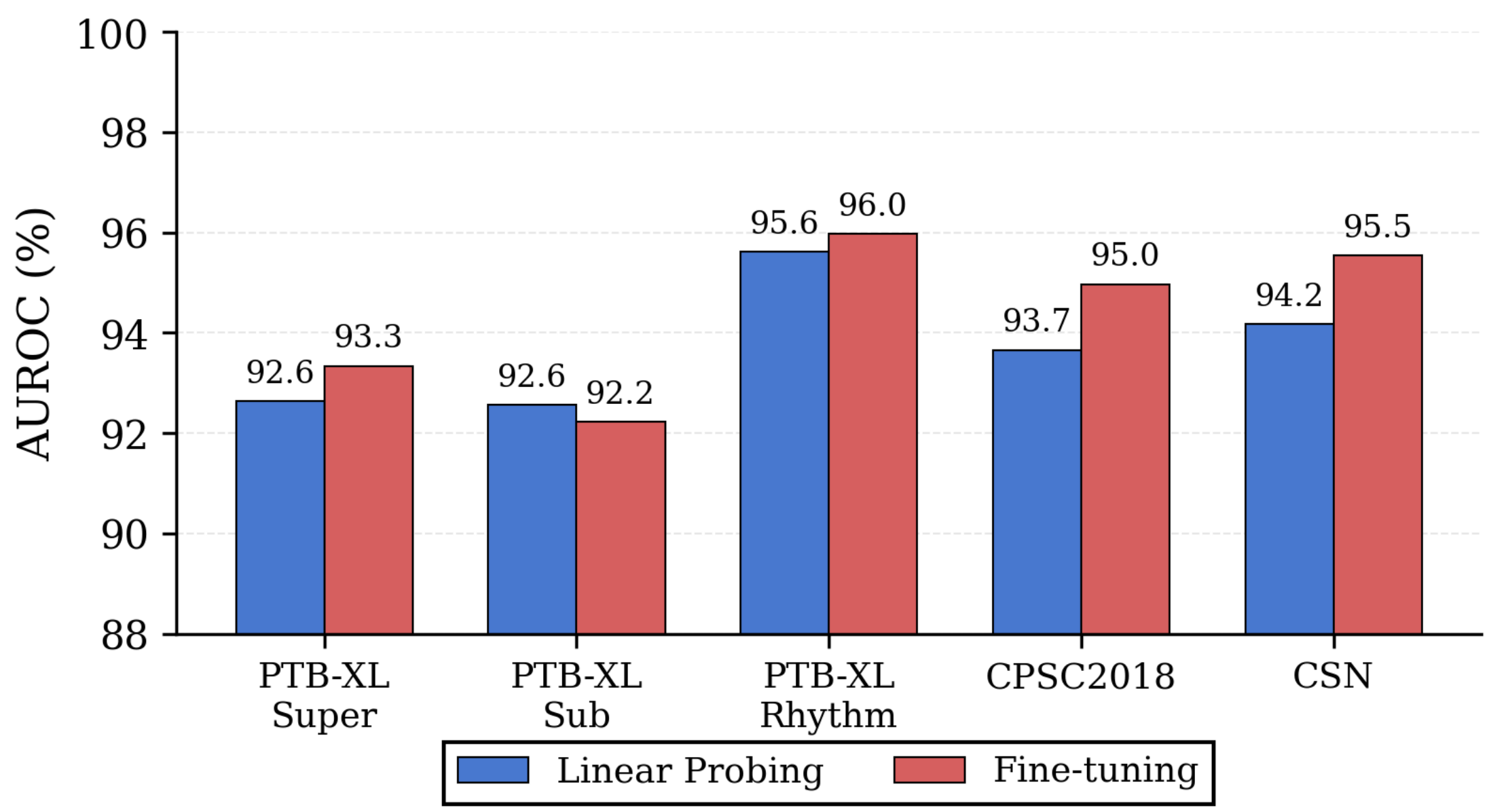

Figure 2. TRACE downstream performance across clinical benchmarks under linear probing and fine-tuning protocols. Error bars indicate standard deviation over five runs.

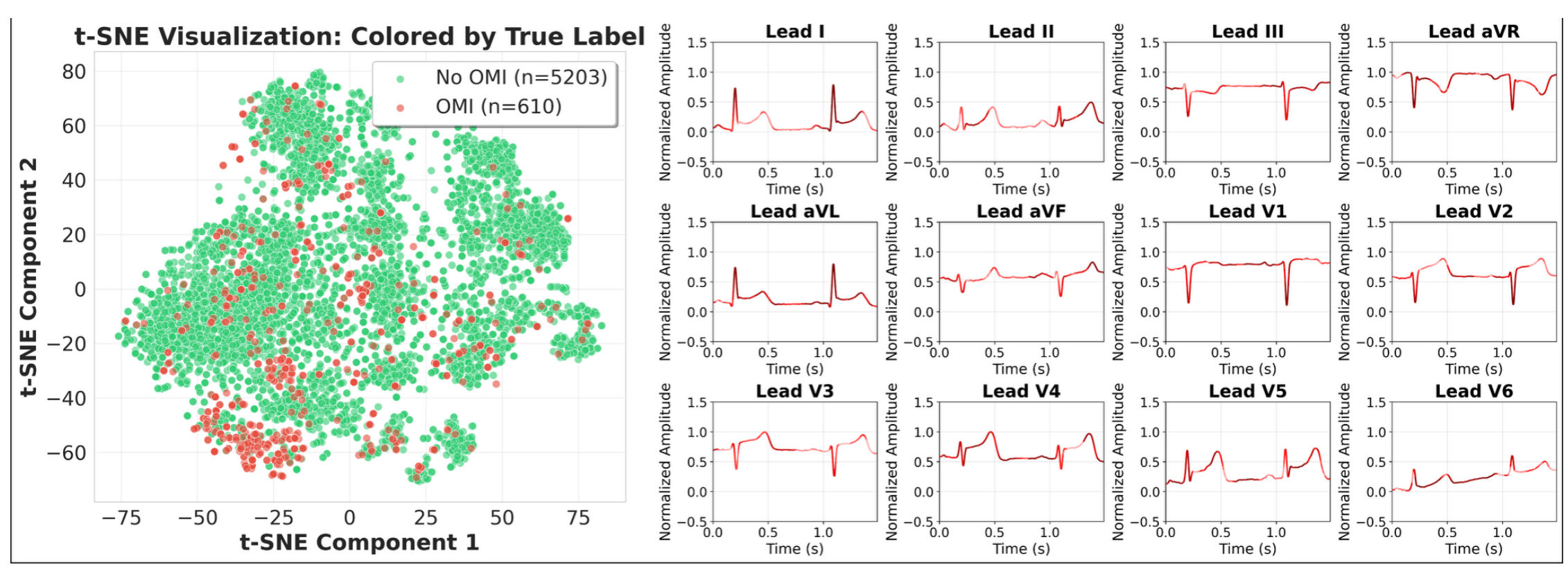

Figure 3. (a). T-SNE plot for TRACE for ACO and non-ACO samples and (b). Saliency map for true positive sample.

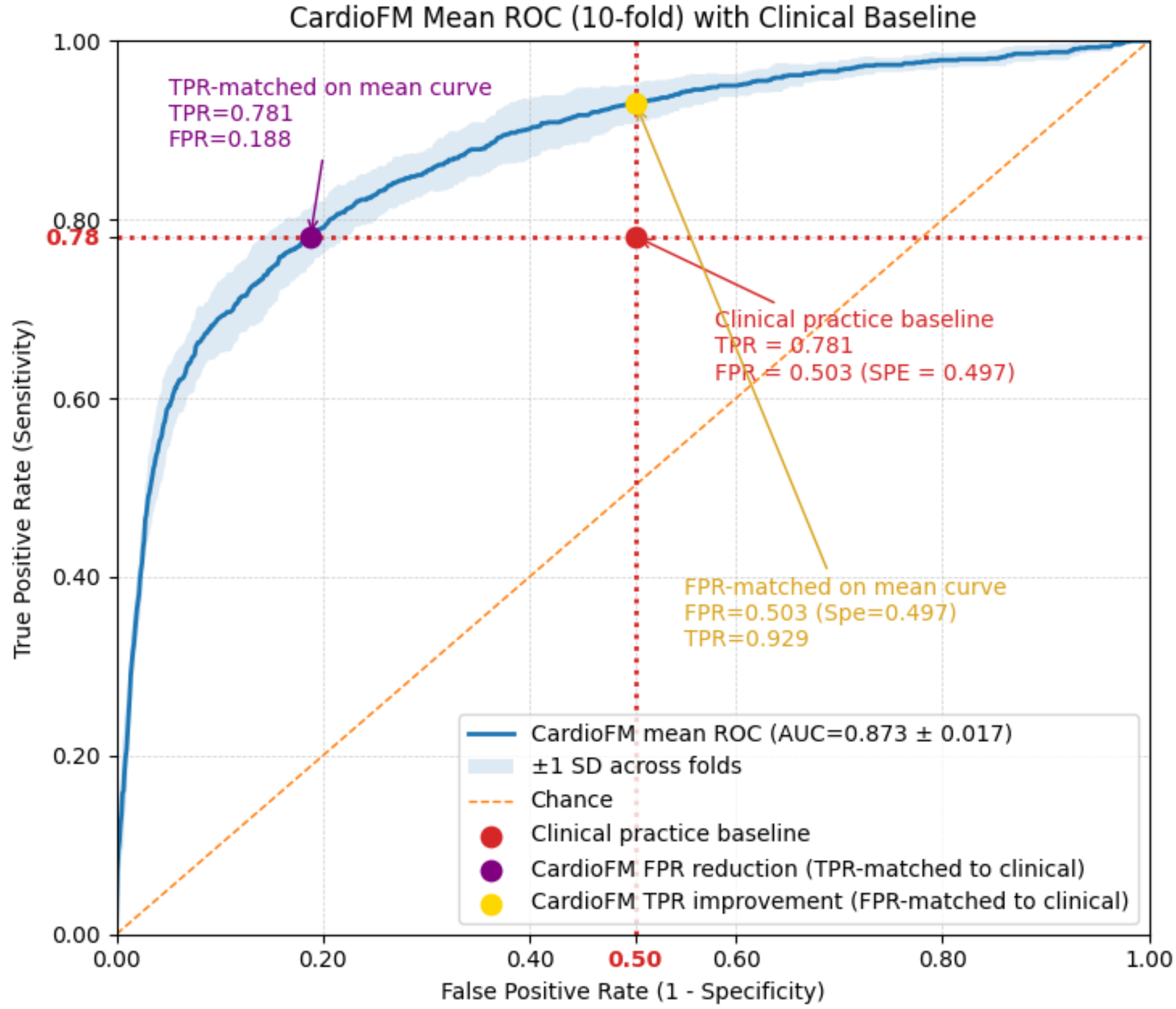

Figure 4. TRACE Performance comparison to clinical practice baseline.

Table 1. Comparison with Multimodal ECG-Text Methods

| Category | Method | Paradigm | PTBXL-Super | PTBXL-Sub | PTBXL-Rhythm | CPSC2018 | CSN |
|---|---|---|---|---|---|---|---|
| Unimodal | SimCLR (Chen et al., 2020) | Contrastive | 73.53 | 73.39 | 77.73 | 76.54 | 73.20 |
| | BYOL (Grill et al., 2020) | Non-Contrastive | 76.45 | 71.64 | 77.17 | 78.75 | 74.69 |
| | BarlowTwins (Zbontar et al., 2021) | Non-Contrastive | 78.41 | 74.34 | 77.62 | 78.39 | 77.43 |
| | MoCo-v3 (Chen et al., 2021) | Contrastive | 78.26 | 76.69 | 74.33 | 75.29 | 77.68 |
| | SimSiam (Chen & He, 2021) | Non-Contrastive | 75.63 | 76.38 | 75.92 | 75.31 | 77.41 |
| | TS-TCC (Eldele et al., 2021) | Contrastive | 78.91 | 77.87 | 78.23 | 78.72 | 76.79 |
| | CLOCS (Kiyasseh et al., 2021) | Contrastive | 76.31 | 76.24 | 76.31 | 77.49 | 76.13 |
| | ASTCL (Wang et al., 2023) | Contrastive | 81.02 | 76.51 | 76.05 | 79.51 | 75.79 |
| | CRT (Zhang et al., 2023a) | Generative | 77.24 | 78.67 | 74.41 | 82.03 | 78.80 |

| Category | Method | Paradigm | PTBXL-Super | PTBXL-Sub | PTBXL-Rhythm | CPSC2018 | CSN |
|---|---|---|---|---|---|---|---|
| Multimodal | ST-MEM (Na et al., 2024) | Generative | 71.36 | 63.59 | 74.85 | 70.39 | 71.36 |
| | ECGFounder (Li et al., 2025) | Supervised | 91.20 | 88.03 | 93.61 | 83.18 | 84.35 |
| | MERL (ResNet) (Liu et al., 2024a) | Contrastive | 88.67 | 84.72 | 88.34 | 90.57 | 87.95 |
| | MERL (ViT) (Liu et al., 2024a) | Contrastive | 85.27 | 82.98 | 81.83 | 89.44 | 84.87 |
| | K-MERL (Liu et al., 2025) | Contrastive | 89.83 | 88.00 | 91.04 | 91.26 | 93.71 |
| | SuPreMe (Cai et al., 2025) | Contrastive | 87.67 | 84.84 | 84.42 | 86.74 | 89.16 |
| | **TRACE (Our Model)** | Hybrid | **92.64** | **92.57** | **95.62** | **93.66** | **94.17** |

## APPENDIX

### A. Downstream Dataset Details

**PTB-XL.** This dataset {Wagner, 2020 #18} includes 21,837 12-lead ECG recordings from 18,885 patients, sampled at 500 Hz with 10-second duration. Three multi-label classification tasks are defined: Superclass (5 categories), Subclass (23 categories), and Rhythm (12 categories). We use the official 70%:10%:20% train:validation:test split.

**CPSC2018.** This dataset {Liu, 2018 #40} contains 6,877 standard 12-lead ECG recordings at 500 Hz sampling rate, spanning 6–60 seconds in duration, annotated with 9 distinct rhythm labels. We use a 70%:10%:20% split.

**Chapman-Shaoxing-Ningbo (CSN).** This dataset {Zheng, 2020 #19} comprises 45,152 12-lead ECG recordings with 11 rhythm annotations, including sinus rhythm, atrial fibrillation, atrial flutter, and various arrhythmias. This benchmark assesses generalization to rhythm disorders distinct from the structural conditions in PTB-XL. We use a 70%:10%:20% split.

**Private ACO Dataset.** The ACO dataset consists of 5,813 prehospital 12-lead ECGs obtained over 6 years from a county-level Emergency Medical Services (EMS) system for patients with suspected acute coronary syndrome. Recordings were linked to receiving hospitals' electronic health records for outcome ascertainment. Ground-truth ACO status was adjudicated by attending cardiologists through comprehensive review of angiographic findings, ECG evolution, cardiac biomarkers, and clinical trajectory. We evaluate using 10-fold cross-validation, with each fold using a 70%:10%:20% train/validation/test split.

### B. LLM Prompt for Report Curation

This provides the complete prompt used in Stage (ii) of our Qwen3-32B-based report curation pipeline. The prompt employs a cardiologist persona with explicit decomposition rules to extract

ECG-specific findings from raw clinical reports while removing noise, uncertainty language, and non-ECG content.

**Prompt Design Principles**

The prompt is designed around three key principles: (1) *role specification* establishes clinical expertise for accurate medical term extraction, (2) *explicit filtering rules* systematically remove uncertainty qualifiers and temporal references that do not reflect the current ECG state, and (3) *few-shot examples* demonstrate the expected transformation from complex clinical narratives to standardized base terms.

**Complete Prompt**

```
You are a cardiologist. Extract ECG findings and decompose them into base clinical terms.

Rules:

1. Remove uncertainty language: "may be", "could be", "possible",
   "suggestive of", "consistent with", "cannot exclude"

2. Remove temporal references: "compared to", "is now", "is no longer",
   "new", "unchanged", "evolving"

3. Remove non-ECG content: medications, imaging findings, patient history,
   clinical recommendations

4. Extract only ECG-attributable observations: rhythm, rate, axis,
   intervals, ST-T changes, conduction abnormalities, morphological features

5. Preserve standard medical terminology without paraphrasing

6. Output as comma-separated lowercase terms


Examples:
```

Input: anterior st segment depression with biphasic to inverted t waves of uncertain significance

Output: st segment depression, anterior, t wave abnormalities, biphasic t waves, t wave inversion

Input: incomplete right bundle-branch block suggestive of right ventricular hypertrophy

Output: incomplete right bundle-branch block, right ventricular hypertrophy

Input: findings raise possibility of prior posterior myocardial infarction

Output: myocardial infarction, posterior

Input: sinus rhythm with first degree av block, unchanged from previous

Output: sinus rhythm, first degree av block

Input: refer to previous ecg for comparison

Output: [EXCLUDE]

Input: patient on metoprolol, ecg shows sinus bradycardia

Output: sinus bradycardia

Now extract findings from the following report:

{report_text}

**Exclusion Criteria**

Reports receiving the [EXCLUDE] output token are removed from the training dataset. This occurs when reports contain only: (1) references to prior ECGs without new findings, (2) administrative notes, (3) medication or imaging information without ECG observations, or (4) illegible or incomplete entries.

**Deployment and Compute**

The entire curation pipeline was run on-premise using locally hosted open-source model weights; no external or commercial API was used at any stage. Qwen3-32B was executed on Emory University's high-performance computing cluster (HyPERC3) using four NVIDIA A100 (80 GB) GPUs. The model was loaded through HuggingFace Transformers with device_map="auto" for automatic multi-GPU sharding across the four A100s and bfloat16 precision to satisfy memory constraints. The 609,272 linked ECG reports were processed in chunked batches of 100 with checkpoint-based resumption for fault tolerance, and the model's thinking mode was enabled to ensure reliable structured extraction. Because the pipeline depends only on open-weight models and standard inference tooling, it can be reproduced without access to any proprietary service.

## C. Clinical Terminology Dictionary

Stage (iii) of our curation pipeline employs a clinical terminology dictionary for normalizing ECG findings to standardized base terms. We adopt the dictionary introduced by {Liu, 2024 #10}, which maps Standard Communications Protocol (SCP) codes to their corresponding clinical term variations. This dictionary was originally developed to enable zero-shot ECG classification by bridging the vocabulary gap between structured diagnostic codes and free-text clinical descriptions. The complete dictionary is available in {Liu, 2024 #10} code repository.

During terminology standardization, abbreviated terms extracted by the LLM (e.g., “RVH”, “RBBB”) are expanded to their full clinical expressions (e.g., “right ventricular hypertrophy”, “right bundle branch block”). This normalization ensures consistent vocabulary across the curated dataset while preserving the clinical semantics required for meaningful ECG-text alignment during contrastive learning.

**D. Ablation Study**

**Component Analysis.** Table A.1 reports ablation results isolating each pretraining component of TRACE. Training with masked autoencoder reconstruction alone (MAE) achieves 0.8623 AUROC on PTBXL-Super under linear probing, confirming that self-supervised signal reconstruction learns morphologically meaningful features. Contrastive multimodal alignment alone (CMA) substantially improves performance to 0.9196 AUROC. It demonstrates that text supervision provides discriminative semantic information by grounding signal patterns in clinical terminology. TRACE’s hybrid CMA+MAE approach outperformes both individual components. This pattern holds consistently across ACO detection and Chapman arrhythmia classification.

**Table A.1.** Performance of each component on pretraining. All values are AUROC (mean ± std). Best results are in bold.

| Method | Evaluation | PTBXL-Super | OMI | CSN |
|---|---|---|---|---|
| MAE | Linear Probing | 0.8623 ± 0.0030 | 0.7965 ± 0.0069 | 0.8146 ± 0.0064 |
| | Fine-tuning | 0.9253 ± 0.0014 | 0.8350 ± 0.0303 | 0.9218 ± 0.0027 |
| CMA | Linear Probing | 0.9196 ± 0.0013 | 0.8490 ± 0.0035 | 0.9173 ± 0.0013 |

| Method | Evaluation | PTBXL-Super | OMI | CSN |
|---|---|---|---|---|
| | Fine-tuning | 0.9253 ± 0.0014 | 0.8558 ± 0.0177 | 0.9459 ± 0.0027 |
| **TRACE (CMA+MAE)** | Linear Probing | **0.9264 ± 0.0011** | **0.8635 ± 0.0102** | **0.9417 ± 0.0054** |
| | Fine-tuning | **0.9334 ± 0.0028** | **0.8731 ± 0.0169** | **0.9555 ± 0.0046** |

The consistent superiority of the hybrid approach demonstrates that reconstruction and alignment objectives are complementary rather than redundant. MAE forces the encoder to capture fine-grained morphological details (QRS morphology, ST-segment deviations) through local patch reconstruction, while CMA provides semantic scaffolding by aligning these features with diagnostic concepts from clinical reports. Joint optimization through uncertainty-weighted multi-task learning enables both objectives to shape the shared encoder simultaneously, avoiding the representational interference that occurs in sequential training where one objective overwrites features learned by the other. Under fine-tuning, TRACE maintains its advantage, indicating that hybrid pretraining already captures most task-relevant features.

**Mask Ratio Analysis.** Table A.2 examines how the MAE masking ratio affects downstream performance across evaluation protocols. TRACE achieves optimal results at 75% masking. Reducing the mask ratio to 70% yields comparable performance, indicating that the encoder benefits from additional visible context for learning local morphological patterns. However, increasing masking beyond 75% causes progressive degradation.

**Table A.2.** Effect of MAE mask ratio on downstream performance. LP: linear probing. FT: fine-tuning. All values are AUROC (mean ± std). Best results are in bold.

| | PTBXL-Super | | PTBXL-Sub | | PTBXL-Rhythm | | OMI | | CSN | | CPSC2018 | |
|---|---|---|---|---|---|---|---|---|---|---|---|---|
| **Mask Ratio** | **LP** | **FT** | **LP** | **FT** | **LP** | **FT** | **LP** | **FT** | **LP** | **FT** | **LP** | **FT** |
| **70%** | 92.19 ± 0.10 | 93.17 ± 0.18 | 92.39 ± 0.68 | **92.27 ± 0.77** | 96.09 ± 0.66 | 96.45 ± 1.80 | 86.08 ± 0.20 | 85.91 ± 1.59 | 93.92 ± 0.42 | **95.76 ± 0.37** | 93.35 ± 0.32 | 94.96 ± 0.20 |
| **75%** | **92.64 ± 0.11** | **93.34 ± 0.28** | **92.57 ± 0.56** | 92.23 ± 0.32 | 95.62 ± 0.65 | 95.98 ± 1.12 | **86.35 ± 1.02** | **87.31 ± 1.69** | **94.17 ± 0.54** | 95.55 ± 0.46 | **93.66 ± 0.14** | **94.97 ± 0.32** |
| **80%** | 91.73 ± 0.23 | 93.14 ± 0.10 | 91.45 ± 0.56 | 92.00 ± 0.50 | 95.76 ± 0.83 | 96.25 ± 1.66 | 82.76 ± 0.56 | 84.66 ± 2.70 | 93.51 ± 0.23 | 95.53 ± 0.30 | 93.12 ± 0.15 | 94.95 ± 0.33 |
| **85%** | 91.69 ± 0.36 | 92.82 ± 0.18 | 90.31 ± 0.73 | 90.86 ± 0.71 | **96.50 ± 0.69** | **96.60 ± 1.97** | 78.33 ± 0.91 | 84.08 ± 1.91 | 93.00 ± 0.50 | 95.27 ± 0.38 | 92.94 ± 0.16 | 94.89 ± 0.38 |
| **90%** | 92.23 ± 0.20 | 92.99 ± 0.20 | 91.54 ± 0.55 | 92.25 ± 0.91 | 95.16 ± 0.55 | 94.93 ± 2.23 | 85.18 ± 0.56 | 86.74 ± 1.68 | 93.67 ± 0.51 | 95.32 ± 0.11 | 92.85 ± 0.39 | 94.49 ± 0.30 |

Interestingly, performance partially recovers at 90% masking on ACO, potentially because extreme masking forces the encoder to rely more on global context and the contrastive alignment objective rather than local reconstruction. However, this recovery does not generalize PTBXL-Super continues to underperform at 90%, and the pattern is inconsistent across other benchmarks. All subsequent experiments use 75% masking as the default configuration.

**Loss Balancing Strategy.** Table A.3 compares approaches for combining reconstruction and contrastive objectives, which operate at fundamentally different scales (MAE: 0.02-0.05, CMA: 4-8). Naive summation (CMA + MAE) allows the larger-magnitude contrastive loss to dominate optimization. Fixed weighting (0.3*CMA + MAE) shows marginal improvement, but requires extensive hyperparameter search and lacks adaptability to changing loss dynamics across training epochs. Dynamic Weight Averaging (DWA), which adjusts weights based on relative loss descent rates, proves unstable when objectives have such disparate magnitudes, indicating that gradient-based balancing mechanisms fail when loss scales differ by two orders of magnitude.

**Table A.3.** Effect of loss balancing strategies on downstream performance. All values are AUROC (%).

| Strategy | Dataset | Linear Probing | Fine-tuning |
|---|---|---|---|
| CMA + MAE | PTBXL-Super | 91.80 ± 0.13 | 92.22 ± 0.20 |

| Strategy | Dataset | Linear Probing | Fine-tuning |
|---|---|---|---|
| | OMI | $83.04 \pm 0.43$ | $83.57 \pm 2.51$ |
| | CSN | $93.36 \pm 0.32$ | $90.82 \pm 0.98$ |
| 0.3*CMA + MAE | PTBXL-Super | $91.92 \pm 0.16$ | $92.23 \pm 0.16$ |
| | OMI | $84.77 \pm 0.55$ | $80.87 \pm 2.67$ |
| | CSN | $93.46 \pm 0.30$ | $91.25 \pm 0.36$ |
| DWA | PTBXL-Super | $89.72 \pm 0.22$ | $92.40 \pm 0.09$ |
| | OMI | $75.49 \pm 2.52$ | $78.28 \pm 2.15$ |
| | CSN | $92.76 \pm 0.77$ | $93.73 \pm 0.64$ |
| **Unc. Weighting** | PTBXL-Super | **$92.64 \pm 0.11$** | **$93.34 \pm 0.28$** |
| | OMI | **$86.35 \pm 1.02$** | **$87.31 \pm 1.69$** |
| | CSN | **$94.17 \pm 0.54$** | **$95.55 \pm 0.46$** |

TRACE's uncertainty weighting consistently outperforms all alternatives. By learning task-specific homoscedastic uncertainty parameters $\sigma_1^2$ and $\sigma_2^2$ that model observation noise, the method automatically calibrates loss contributions through precision terms $1/(2\sigma_i^2)$ tasks with higher intrinsic uncertainty receive lower weight during optimization. This principled probabilistic framework eliminates manual tuning while adapting to changing task difficulties throughout training, enabling stable joint optimization despite the scale mismatch. The gains persist under fine-tuning, confirming that uncertainty weighting produces more robust pretrained representations compared to heuristic balancing schemes.

**Text Encoder Selection.** Table A.4 compares text encoders for the contrastive alignment branch of TRACE under the final uncertainty-weighted configuration. We evaluate our domain-specific encoder MedCPT a retrieval model contrastively pretrained to align short biomedical queries with relevant passages against Qwen3-32B, a state-of-the-art open-weights LLM nearly $300\times$ larger, whose frozen text embeddings we substitute for MedCPT while retraining TRACE identically. MedCPT outperforms Qwen3-32B on every task under linear probing, by 2.6–4.6 AUROC points (mean 3.7), and on five of six tasks under fine-tuning; Qwen3-32B surpasses MedCPT only on CPSC2018 fine-tuning. The gap is largest under linear probing which most directly reflects the quality of the aligned representation and narrows under fine-tuning as the ECG encoder adapts to each task. We attribute this to a mismatch of inductive bias rather than scale: MedCPT's contrastive, similarity-structured embedding space mirrors the ECG–report alignment objective, whereas a decoder-only generative model such as Qwen3-32B is not optimized to produce embeddings whose cosine geometry supports InfoNCE-style learning. Model recency and parameter count alone therefore do not translate into better ECG–text supervision. This is consistent with our broader domain strategy: we deploy Qwen3-32B where generative reasoning is advantageous curating and decomposing free-text reports (Report Curation Pipeline) while retaining a retrieval-contrastive encoder for the alignment objective itself. In a preliminary ablation, PubMedBERT the masked-language model MedCPT is built upon performed comparably to MedCPT and slightly below it under linear probing, further supporting the use of a retrieval-pretrained rather than a generic biomedical encoder.

**Table A.4.** Effect of text encoder choice on downstream performance, comparing our domain-specific MedCPT encoder against a state-of-the-art general-purpose LLM (Qwen3-32B) as the text backbone, under the final TRACE configuration (uncertainty-weighted CMA+MAE). All values are AUROC (mean ± std over 5 seeds); best per task and protocol in bold. LP: linear probing. FT: fine-tuning.

| Dataset | MedCPT (LP) | MedCPT (FT) | Qwen3-32B (LP) | Qwen3-32B (FT) |
|---|---|---|---|---|
| PTBXL-Super | **0.9264 ± 0.0011** | **0.9334 ± 0.0028** | 0.9001 ± 0.0021 | 0.9284 ± 0.0023 |
| PTBXL-Sub | **0.9257 ± 0.0056** | **0.9223 ± 0.0032** | 0.8883 ± 0.0130 | 0.9204 ± 0.0103 |
| PTBXL-Rhythm | **0.9562 ± 0.0065** | **0.9598 ± 0.0112** | 0.9229 ± 0.0172 | 0.9571 ± 0.0297 |
| CPSC2018 | **0.9366 ± 0.0014** | 0.9497 ± 0.0032 | 0.8902 ± 0.0080 | **0.9581 ± 0.0026** |
| CSN | **0.9417 ± 0.0054** | **0.9555 ± 0.0046** | 0.9083 ± 0.0058 | 0.9460 ± 0.0041 |
| OMI | **0.8635 ± 0.0102** | **0.8731 ± 0.0169** | 0.8210 ± 0.0082 | 0.8586 ± 0.0147 |